\documentclass[letterpaper, 10 pt, conference]{ieeeconf}  

\IEEEoverridecommandlockouts                              

\usepackage{amsmath}
\usepackage{amssymb}
\usepackage{bm}
\usepackage{graphicx}
\usepackage{subcaption}
\usepackage{hyperref}
\usepackage{stackengine}
\usepackage{bbm}
\usepackage{tabularx}
\usepackage{dblfloatfix}
\usepackage{flushend}

\usepackage{dsfont}
\usepackage{tikz}
\usetikzlibrary{shapes.geometric}
\usetikzlibrary{arrows.meta}
\usetikzlibrary{calc}
\usetikzlibrary{positioning}
\usetikzlibrary{decorations.pathreplacing}
\usetikzlibrary{tikzmark}

\usepackage{booktabs}
\usepackage{cite}

\usepackage{layouts}
\usepackage{units}
\usepackage{nicefrac, xfrac}
\usepackage{multirow}
\usepackage{array}
\newcolumntype{M}[1]{>{\centering\arraybackslash}m{#1}}

\definecolor{teacher_green}{rgb}{0.55, 0.8, 0.7} 
\definecolor{student_orange}{rgb}{0.802, 0.492, 0.627} 

\title{\LARGE \bf Prompt-responsive Object Retrieval with \\Memory-augmented Student-Teacher Learning}

\author{
Malte Mosbach and Sven Behnke\\%
\thanks{All authors are with the Autonomous Intelligent Systems Group, University of Bonn, Germany and the Lamarr Institute for ML and AI, Bonn, Germany.
The corresponding author is Malte Mosbach {\tt\small mosbach@ais.uni-bonn.de}}%
}

\begin{document}
\maketitle

\begin{abstract}
  Building models responsive to input prompts represents a transformative shift in machine learning.
  This paradigm holds significant potential for robotics problems, such as targeted manipulation amidst clutter.
  In this work, we present a novel approach to combine promptable foundation models with reinforcement learning (RL), enabling robots to perform dexterous manipulation tasks in a prompt-responsive manner.
  Existing methods struggle to link high-level commands with fine-grained dexterous control.
  We address this gap with a memory-augmented student-teacher learning framework.
  We use the Segment-Anything 2 (SAM\,2) model as a perception backbone to infer an object of interest from user prompts.
  While detections are imperfect, their temporal sequence provides rich information for implicit state estimation by memory-augmented models.
  Our approach successfully learns prompt-responsive policies, demonstrated in picking objects from cluttered scenes.
  Videos and code are available at
  \url{https://memory-student-teacher.github.io}
\end{abstract}


  \section{Introduction}

  Foundation Models (FMs) such as GPT-4~\cite{GPT-4} and Segment Anything~\cite{Kirillov2023} represent a paradigm shift in the field of artificial intelligence.
  Trained on broad web-scale datasets, these models excel in generating contextually nuanced outputs across a diverse array of tasks~\cite{Zitkovich2023}.
  This capability is typically implemented through prompt engineering, where human understandable inputs are used to prompt the model for a valid response to the task at hand~\cite{Kirillov2023, Brown2020}.
  Thus, simple instructions can be used to condition a model to perform a myriad of downstream tasks.

  Being able to control the behavior of dexterous robots in a similar manner is a long-standing goal~\cite{Parakh2023}, yet existing approaches mainly leverage FMs for high-level planning~\cite{Parakh2023, Huang2022, Ahn2022}..
  While this approach has yielded impressive capabilities in terms of developing versatile agents, it falls short in replicating the intricate, low-level dexterity required for complex manipulation tasks, such as dexterous grasping from clutter.
  It remains unclear how such approaches can scale to match the intuitive dexterity humans exhibit — often relying on tacit, hard-to-describe skills.~\cite{Dreyfus1987}.

  In contrast, reinforcement learning (RL) bypasses the need for explicit, high-level instructions, opting instead to learn behaviors through trial-and-error. Recent works demonstrate that RL is capable of learning fine motor behaviors comparable to human dexterity~\cite{Chen2022,Chen2022a,Petrenko2023,Ma2023}, yet the learned policies tend to be task-specific and lack the adaptability of prompt-conditioned models.

  Consider the scenario of a warehouse robot tasked with fulfilling an order by picking specific items from cluttered bins.
  Retraining for each new object or providing explicit models for every item to grasp is neither practical nor scalable.
  Instead, we aim to operate such a system using abstract, human-understandable instructions, like a packing list of items to retrieve.
  However, existing methods fail to integrate intuitive, language-guided instructions with the low-level control needed for dexterous manipulation.

  \begin{figure}[t]
	\centering
	\includegraphics[width=\linewidth]{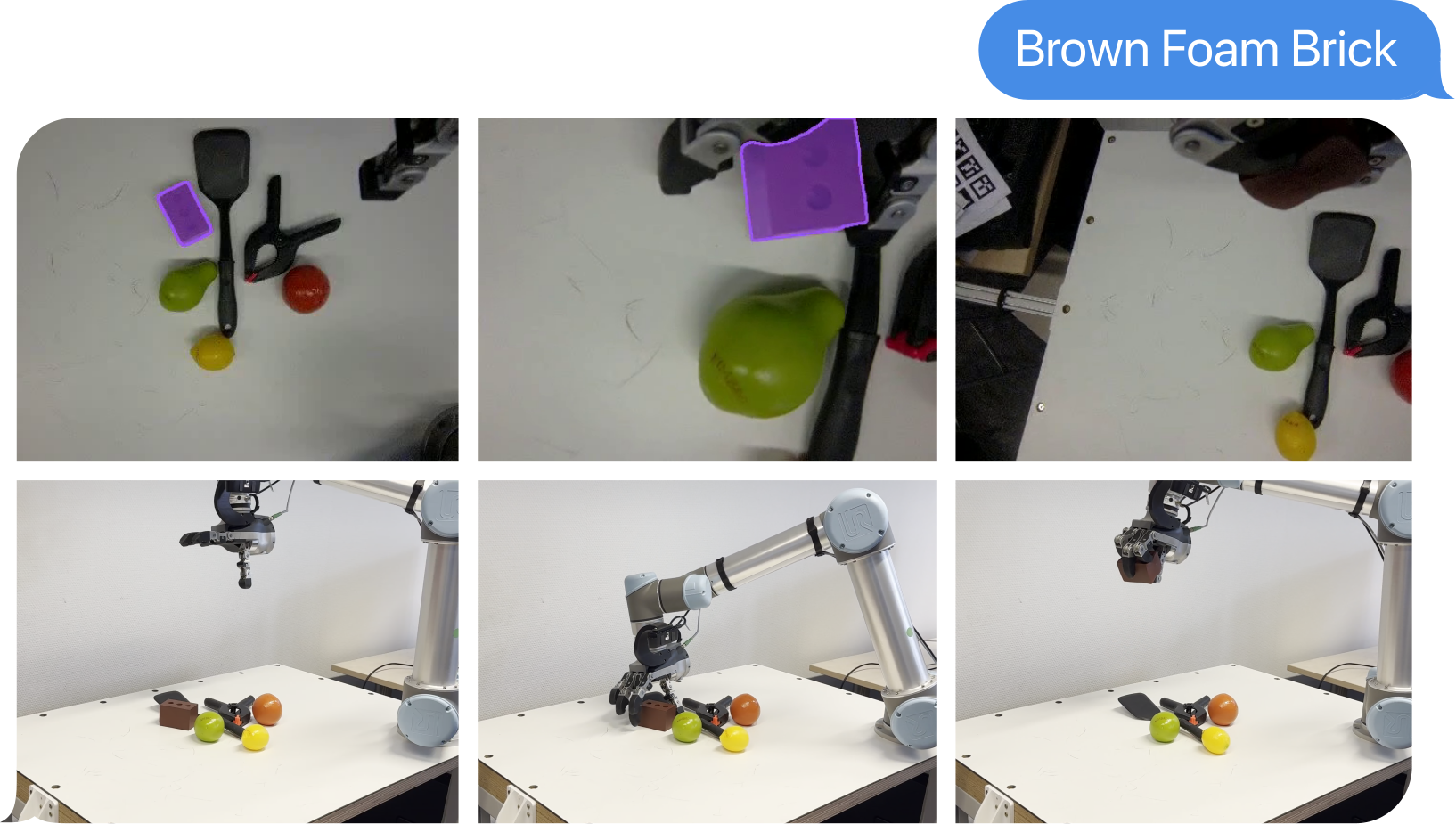}
	\caption{We present a student-teacher framework that enables learning interactive, prompt-responsive policies for object retrieval from cluttered scenes (Figure inspired by~\cite{Shen2023}).}
	\label{fig:teaser}
  \end{figure}

  To bridge this gap, we propose a novel approach that integrates the broad, open-vocabulary capabilities of vision foundation models (VFMs) with the precise motor control developed through RL.  
  Specifically, we leverage the Segment Anything 2 (SAM\,2)~\cite{Ravi2024} model as a perception backbone to segment objects of interest based on user prompts. While the representations generated by SAM\,2 are inherently imperfect -- prone to issues like occlusions or unstable segmentations -- the sequence of outputs over an episode is highly informative and allows for an implicit estimation of the true object state.

  This type of implicit inference of an underlying state from impaired perception via memory-augmented agents has recently been shown to be a powerful tool in navigation of simulated agents~\cite{Wijmans2023} and quadruped robots~\cite{Zhang2024}.  
  While prior works utilized synthetic simulations of imperfect perception, allowing them to train RL agents with high performance, having the SAM\,2 model in the RL training loop pushes the compute requirements beyond what is currently feasible. Instead, we factor out learning how to act and learning to infer the underlying state of an object of interest via a student-teacher formulation. After a teacher policy has been trained to master the task from privileged, simulator-state information, we distill its knowledge to a memory-augmented student policy that operates based on imperfect outputs from our VFM. This distillation process forces the student to implicitly learn the true state of the object to imitate the teacher's actions, even when faced with incomplete or faulty perception data. We explore both LSTMs and Transformers as sequence processing modules, to understand the impact of history-awareness on the student's performance.

  Our key insight is that while SAM\,2 does not allow for reconstructions of Markovian, ground-truth states directly that are needed to deploy a privileged policy, learning to match the teacher's actions forces the student to implicitly learn the associations between the detection sequence and the current state of the object of interest.

  Our results demonstrate that this approach successfully learns dexterous, prompt-responsive policies capable of generating complex, targeted manipulation in cluttered environments.
  Further, the observation-space of the student policy allows for zero-shot real-robot transfer.
  By harnessing the synergy between high-level instruction and low-level robot action, transforming the warehouse robot into a system that picks recyclable cans from trash or selectively removes rotten fruits from a box of produce can be achieved simply through conditioning on human-understandable prompts.\\
  In summary, we address the following research questions:
  \begin{enumerate}
    \item \textit{Can VFMs be used as perception backbones for prompt-responsive manipulation policies?} Yes, we find that the proposed method yields agents that are highly effective at grasping a wide variety of objects on the basis of human-understandable prompts.
    \item \textit{What are the necessary mechanisms for this approach?} The two core mechanisms are first, keeping the FM out of the RL loop via student-teacher learning, and second, using history-aware architectures to implicitly infer underlying states from imperfect detections.
    \item \textit{Does this method transfer to real robot systems?} Yes, we demonstrate that the strong performance of the policies in simulation transfers to our real robot system.  
  \end{enumerate}

  \section{Related Work}
  \subsection{Learning-based Robotic Grasping}
  Reliable robotic grasping has been a prominent challenge in robotics for decades~\cite{Xie2023}. The control aspect of this problem has been formulated mainly as either (1) a problem of grasp-pose prediction or (2) as closed-loop continuous control.

  Grasp-pose prediction is typically solved through supervised learning, where a model is trained to predict the best grasp pose for a given object. Redmon and Angelova~\cite{Redmon2015} introduced a deep learning approach to predict grasp positions on objects using convolutional neural networks.

  The latter group of continuous-control approaches typically utilize RL. Levine et al.~\cite{Levine2016} demonstrated the use of deep RL for end-to-end training of robotic grasping policies. More recently, Kalashnikov et al.~\cite{Kalashnikov2018} developed QT-Opt, a scalable RL algorithm that significantly improves grasping performance. Mosbach and Behnke~\cite{Mosbach2024} utilize SAM for prompt-based robotic grasping. However, they consider only tabletop grasping and rely on unobstructed tracking of the object throughout the episode. Our approach eliminates this requirement, allowing target objects to go out of view, be occluded, or be misdetected by the model.

  \subsection{Learning from Privileged Information}
  Chen et al.~\cite{Chen2020} observed that learning to imitate expert drivers from visual perception conflates two difficult problems: learning to perceive the environment and learning to control the vehicle. They proposed a two-stage approach where a teacher policy is trained to imitate the expert's actions based on the environment's ground-truth state. The knowledge from the teacher is then distilled into a vision-based student policy.
  Recently, Chen et al.~\cite{Chen2022, Chen2022a} extended this strategy to reinforcement learning. In their approach, a teacher policy is trained using simulator state information to solve the control problem. Subsequently, a visual student policy is trained more efficiently in a supervised manner to imitate the teacher from realistic observations.
  Kumar et al.~\cite{Kumar2021} identified domain randomization parameters as a special kind of privileged information. Their method, Rapid Motor Adaptation, enables a student policy to infer domain parameters from observation history, which are made available to the teacher policy.
  Building on this framework, Margolis et al.~\cite{Margolis2024} developed a method to learn agile locomotion over diverse terrains. The domain randomization parameters are made available to the teacher policy, to avoid learning overly conservative behaviors. The student policy, which cannot access these parameters directly, utilizes a history-aware approach, $\pi_{S}(\bm{x}_t, \bm{x}_{[t-h:t-1]})$, to implicitly deduce the domain parameters from the observation history.
  Zhang et al.~\cite{Zhang2024} recently utilized an asymmetric actor-critic architecture to achieve resilient navigation. In their method, the critic has perfect observability of obstacles, while the history-aware actor uses exteroception along with an imperfect map for navigation. This approach enables policies to learn about unknown obstacles by integrating their history of observations from interactions such as bumping into or touching them.
  Similar to our approach, Kumar et al. and Margolis et al.~\cite{Kumar2021, Margolis2024} use student-teacher learning to overcome a problem of explicit versus implicit observability. While their focus is on domain randomization, we tackle the problem associating imperfect sequences of detections with the underlying state of the scene.

  \begin{figure*}[t]
	\centering
	\begin{subfigure}{1.0\textwidth}
		\centering
		\includegraphics[width=\linewidth]{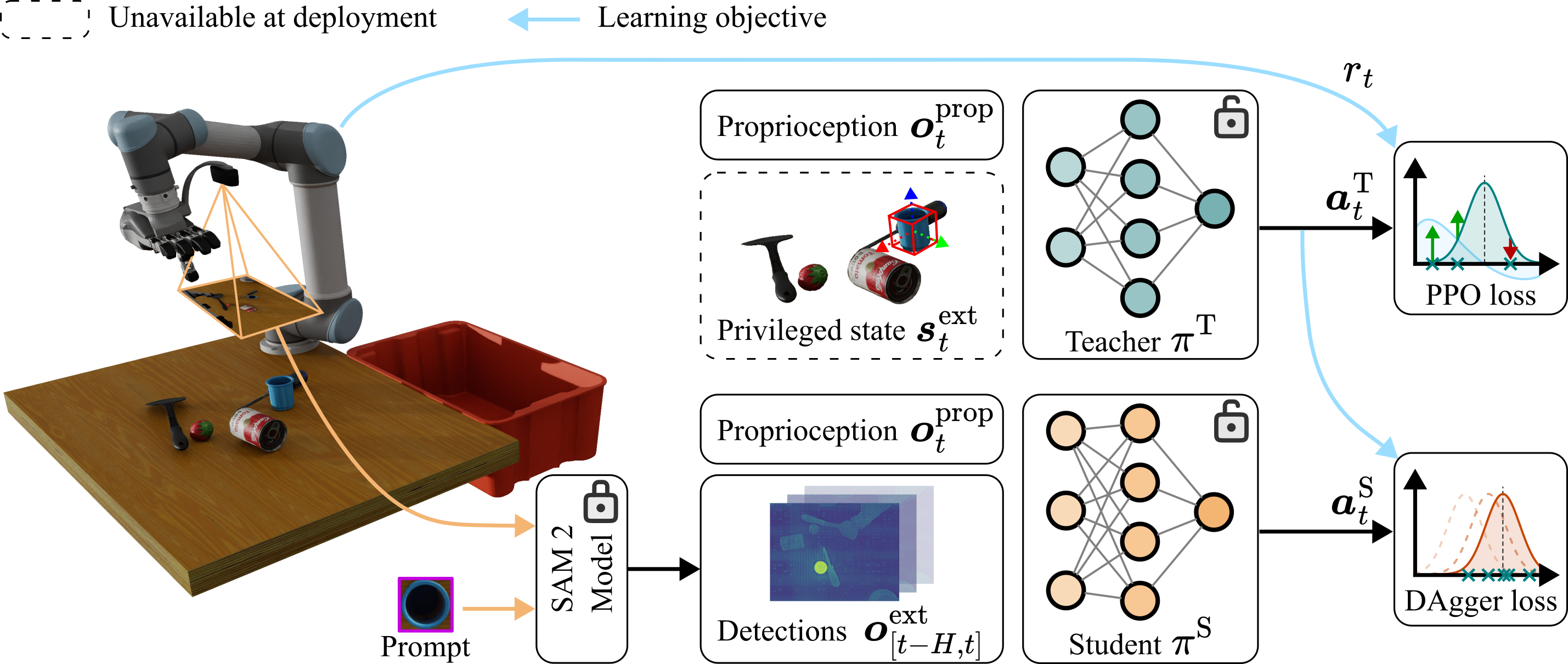}
		\caption{Training in simulation is performed in two stages via student-teacher learning.}
	\end{subfigure}

	\vspace{0.5cm}

	\begin{subfigure}{1.0\textwidth}
		\centering
		\includegraphics[width=\linewidth]{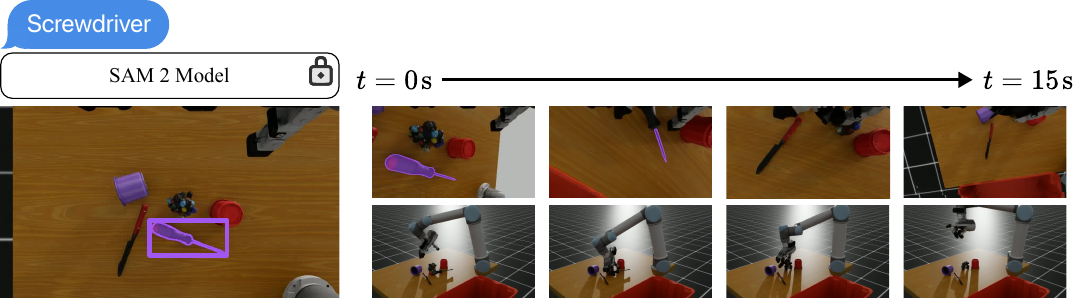}

	\caption{At deployment, the student policy can be prompted with points or language to infer the target object.}
	\end{subfigure}
  
	\caption{We propose to train prompt-guided policies in two stages. First, the teacher policy is trained with model-free RL to solve the control problem from privileged information $\bm{s}_t^\mathrm{ext}$. Thereafter, the student policy is trained to imitate the teacher without access to $\bm{s}_t^\mathrm{ext}$, forcing it to implicitly infer the object state from the history of visual observations $\bm{o}_{[t-H, t]}^\mathrm{ext}$.}
	\label{fig:overview}
  \end{figure*}

  \subsection{Prompt-Guided Manipulation}
  Recent research efforts strive to transfer grounded world knowledge from vision-language models to robotics by training large foundation models on a wide array of behavioral data~\cite{Brohan2023, Brohan2022,Zitkovich2023}.
  Additionally, language models have been employed for high-level planning in robotic manipulation~\cite{Parakh2023, Brohan2023}.
  Recently, Shen et al.~\cite{Shen2023} distilled knowledge from language-supervised image models into a 3D representation of the scene, enabling the learning of 6-DoF grasping of novel objects in a language-guided manner from only a few demonstrations.
  In contrast, our work does not rely on learning a representation of one specific scene beforehand.

  \section{Method}
  \label{sec:method}

  \subsection{Overview}
  Consider a warehouse robot tasked with fulfilling an order by retrieving specific items from cluttered bins.
  At the start of each trial, the robot receives a description of the target object, provided as an open-vocabulary text prompt, a selected point in the camera image, or a detected bounding box.
  The robot must use this instruction to locate and grasp the specified object.

  Successfully executing this task requires addressing two key challenges:
  (1) Interpreting the provided instruction in conjunction with visual observations, and
  (2) mastering the dexterous control needed to grasp objects in cluttered scenes.
  Rather than tackling both challenges simultaneously, we follow recent works in imitation and reinforcement learning~\cite{Chen2020,Chen2022,Chen2022a} and decompose the problem into two stages.

  First, we train a teacher policy using privileged simulator-state information, allowing it to acquire effective control strategies in an idealized setting.
  We then transfer these behaviors to a student policy that operates solely on real-world inputs.
  To bridge the gap between noisy perception and reliable control, we employ a memory-augmented student-teacher learning framework, where the student integrates outputs from perception modules (such as SAM\,2) — despite occlusions and segmentation errors — into an implicit understanding of the object state.
  This approach enables the policy to maintain robustness under imperfect perception while allowing prompt-based object retrieval in cluttered environments.
  
\begin{table*}[!b]
    \begin{minipage}{.45\linewidth}
      \footnotesize
      \caption{Observations combine robot proprioception with \textcolor{teal}{privileged} or \textcolor{orange}{VFM-based} exteroception for the teacher and student, respectively.} \centering
      \label{tab:observations}
      \begin{tabular}{lr}
        \toprule
        Term & Dimensionality \\
        \midrule
        Last actions & $11D$ \\
        Arm joint state & $18D$ \\
        Hand joint targets & $11D$ \\
        Goal position & $3D$ \\

        \addlinespace[0.125cm]

        \tikzmarknode{teacher_start}{Fingertip poses} & $35D$ \\
        Fingertip velocities & $30D$ \\
        Target object OBB corners & $24D$ \\
        Heightmap & $64D$ \\
        Target object velocity & $6D$ \\
        Target to goal pos& \tikzmarknode{teacher_end}{$3D$} \\

        \addlinespace[0.175cm]
        
        \tikzmarknode{student_start}{SAM2 detected point-cloud} & \phantom{A} \tikzmarknode{student_end}{$4D * N_{points}$} \\

        \addlinespace[0.075cm]
        \bottomrule
\end{tabular}

\begin{tikzpicture}[overlay, remember picture, line width=1pt, rounded corners=2pt, line cap=round]
  \draw[dashed, teal] ($(teacher_start.north west)+(-1mm,1mm)$) rectangle ($(teacher_end.south east)+(1mm,-1mm)$);
  \draw[dashed, orange] ($(student_start.north west)+(-1mm,1mm)$) rectangle ($(student_end.south east)+(1mm,-1mm)$);

  \end{tikzpicture}

    \end{minipage}%
    \hspace{0.05\linewidth}%
    \begin{minipage}{.5\linewidth}
    \footnotesize
    \caption{Reward terms used to train the teacher policy.}\centering
    \vspace*{-1ex}
    \label{tab:rewards}
    \begin{tabular}{llr}
      \toprule
      Term & Equation & Weight \\
      \midrule
      Alive & 1.0 & 0.01 \\
      Grab object & $-\Delta d^\mathrm{grab}$ & 10.0 \\
      Lift object & $\min(h_t, h^\mathrm{lifted})$ & 40.0 \\
      Reach goal & $-\Delta d^\mathrm{goal}$ & 100.0 \\
      Goal bonus & $\mathds{1}(d_t^\mathrm{goal} < \bar{d}^\mathrm{goal})$ & 10.0 \\
      \bottomrule
\end{tabular}
\label{tab:reward_terms}

\vspace{0.125cm}
\vspace*{2ex}

\caption{Termination conditions, where $\mathcal{A}$ and $\mathcal{T}$ denote the bodies of the robot arm and the table (and bin), respectively.}\centering
\vspace*{-1ex}
    \begin{tabular}{llr}
      \toprule
      Term & Condition &  \\
      \midrule
      Arm contacts & $\max_{i \in \mathcal{A}} \|\bm{c}_{t}^i\|_2 > 5.0$ \\
      Tabletop or bin contacts & $\max_{i \in \mathcal{T}} \|\bm{c}_{t}^i\|_2 > 25.0$ \\
      Time-out & $t > T_{max}$ \\
      \bottomrule
\end{tabular}
\label{tab:termination conditions}

    \end{minipage}
    \end{table*}

  \subsection{Observations}
  To train our prompt-responsive grasping policies, we utilize three types of observations:
  proprioception, privileged exteroception, and VFM-based exteroception (see Table \ref{tab:observations}).

  \textbf{Proprioception:}
  Proprioception encompasses information that is available in both simulation and real-world deployment.
  This includes the joint states of the robotic arm and hand, the most recent action taken, and the 3D goal position for the target object.

  \textbf{Privileged Exteroception:}
  During privileged training, the teacher policy has access to additional information unavailable in real-world deployment.
  This includes the oriented bounding box (OBB) of the target object, as well as a privileged heightmap centered around the gripper, which provides a structured representation of the surrounding clutter.
  Moreover, we provide additional information about the state of the manipulator including the fingertip poses and velocities.

  \textbf{VFM-based Exteroception:}
  The student policy receives object detections from a VFM to provide a prompt-responsive visual input space.
  Since these detections are imperfect and non-Markovian — due to occlusions and misdetections — we provide the student model with a history of recent detections, allowing it to infer the true object state over time.

  \begin{figure*}[t]
    \centering
      \includegraphics[width=\linewidth]{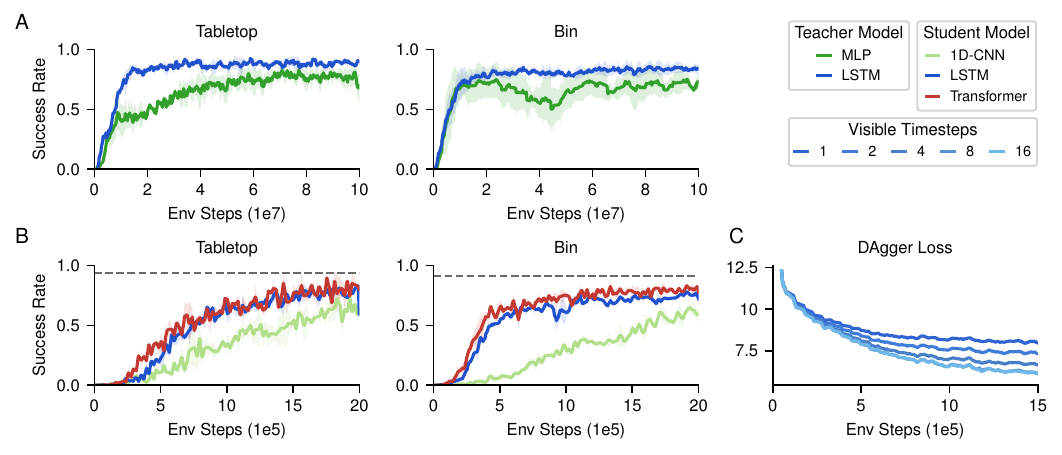}
      \vspace{-0.5cm}
      \caption{Training performance of \text{(A)} the teacher policies, \text{(B)} the student policies, and \text{(C)} the imitation loss over time steps.
      The dashed horizontal lines indicate the performance of the teacher policy used by DAgger.}
	\label{fig:results}
  \end{figure*}

  \subsection{Learning from Privileged Information}
  We train the teacher policy $\pi^\mathrm{T}$ using RL, where the agent maps observations $\bm{o}_t^\mathrm{T}$ to actions $\bm{a}_t$.
  To ensure generalization across diverse objects, we design the simulation as a multi-task learning problem, where each parallel environment contains a randomly selected subset of training objects.
  As a result, the teacher policy must learn to handle objects of varying shapes and sizes as targets and obstacles.
  We use PPO~\cite{Schulman2017} to optimize the teacher policy as it has been shown to perform robustly on difficult continuous control tasks.

  Although the agent has access to privileged simulator-state information, the observation $\bm{o}_t^\mathrm{T}$ at a single time-step $t$ does not convey the full state information, such as the exact shape of an object. 
  Hence, we evaluate the use of LSTM architectures~\cite{Hochreiter1997} alongside a standard MLP policy to enable the teacher policy to consider temporal dependencies for decision-making. The MLP policy comprises three layers with 768, 512, and 256 units, respectively. The LSTM variant adds a single LSTM layer with 768 units before the MLP.

  The teacher policy observes regular proprioception alongside the privileged simulator-state information (see Table \ref{tab:observations}).
  The policy controls the robot's joints at a frequency of 10\,Hz. We use an exponential moving average (EMA) to control the joint velocities of the arm, formulated as $\bm{\dot{q}}^{\text{target}}_{t+1} = \alpha \bm{a}_t + (1 - \alpha) \bm{\dot{q}}^{\text{target}}_{t}$, balancing smoothness and responsiveness.
  The Schunk SIH hand is controlled via servo-actuated tendons.
  A similar EMA formulation is used to set the servo target positions as $\bm{p}^{\text{target}}_{t+1} = \alpha \bm{a}_t + (1 - \alpha) \bm{p}^{\text{target}}_{t}$.

  \textbf{Rewards and Termination Conditions:}
  The reward function is designed to facilitate directed exploration without distracting from the overall objective. Initially, the agent is motivated to move its hand closer to the target object, where $\Delta d^\mathrm{grab}$ denote the change in distance between the fingertips and the object of interest. Once the agent reaches the object, this reward term is exhausted, allowing the agent to focus on manipulating the object. At this stage, we reward the agent for lifting the object from the table or bin, where $h_t$ represents the height of the object as measured by the lowest point of its OBB. Finally, the agent is rewarded for moving the target object to the goal position and for reaching the goal position within a small threshold. The detailed reward function is shown in Table~\ref{tab:reward_terms}.

  Notably, we opted not to include reward terms that directly encourage safe behavior, such as contact or action penalties, as they can interfere with task-relevant exploration in hand-arm manipulation tasks.
  Our experiments showed that imposing penalties for hard contacts significantly hindered effective exploration.
  Instead, we employ termination conditions to enforce safety, which offers several advantages.  Firstly, this eliminates the trade-off between task reward and safety penalties, ensuring that unsafe behaviors are prohibited outright. Secondly, using terminations to punish unsafe behaviors creates a natural learning curriculum. Initially, when the agent expects to obtain low rewards, terminating an episode due to collision is less significant. As the agent improves and anticipates higher rewards, it will avoid these terminations due to the higher cost of lost future rewards. 
  The termination conditions are shown in Table~\ref{tab:termination conditions}.

  \subsection{Memory-augmented Student-Teacher Learning}
  We apply a memory-augmented student-teacher learning framework to transfer knowledge from the trained teacher policy to a student policy that operates solely on proprioception and detections from a VFM.
  SAM\,2 understands the relationship between user prompts and objects.
  Using its detections as a prompt-responsive observation space removes the need for the agent to learn object-prompt associations explicitly.
  Instead, the student policy learns to interpret these detections as part of its perception pipeline.

  The key challenge is that SAM\,2's detections are non-Markovian — occlusions and segmentation inconsistencies can cause missing or unstable detections.
  Thus, for the student policy to successfully imitate the teacher, it must implicitly infer the true state of the target object by integrating past detections with object dynamics.
  This motivates the use of history-aware architectures, which allow the student to compensate for gaps in perception.

  While our final policy should be controllable via user prompts, training requires automated prompting to ensure efficiency and consistency across parallelized simulation environments.
  To achieve this, we introduce two key extensions.
  First, we modify SAM\,2 to handle batched image streams, enabling efficient processing across multiple parallel environments.
  This allows detections to be generated in parallel rather than sequentially, significantly reducing computational overhead.
  Second, we automate prompt generation by leveraging ground-truth object geometry from simulation.
  Specifically, we extract points on the mesh surface of the target object and project them into the camera frame.
  From these projected points, we compute a tight bounding box in the image space, which is then used as the prompt for SAM\,2, ensuring alignment between the model's detections and the true target object defined in the task.

  To process these detections, we transform the depth image from the RGB-D camera into a point cloud, providing a direct 3D representation of objects in the environment.
  Additionally, we append a feature dimension that indicates which subset of points SAM\,2 assigns to the target object.
  To encode these point clouds, we employ a PointNet-like encoder, which extracts an embedding of the scene.
  This embedding is then concatenated with proprioceptive observations and passed to the policy network.
  To integrate temporal information, we evaluate the follwoing memory-based student architectures within the DAgger imitation learning framework~\cite{RossGB:AISTATS11DAgger}.  

  \begin{table*}[t]
    \begin{minipage}{.59\linewidth}
        \footnotesize
        \caption{Success rates for lifting the object and reaching the goal.}\centering
        \label{tab:simulation_results}
        \begin{tabular}{llccccc}
          \toprule
          \multirow{3}{*}{State} & \multirow{3}{*}{Model} & \multicolumn{2}{c}{Tabletop} & & \multicolumn{2}{c}{Bin} \\
          \cmidrule{3-4} \cmidrule{6-7} & & $\textrm{lifted}$ & $\textrm{goal}$ & & $\textrm{lifted}$ & $\textrm{goal}$ \\
          \midrule
          \multirow{2}{*}{Privileged} & MLP & $89.1 \scriptsize{\raisebox{1pt}{$\pm 3.3$}} $ & $86.0 \scriptsize{\raisebox{1pt}{$\pm 2.8$}} $ & & $84.5 \scriptsize{\raisebox{1pt}{$\pm 2.3$}} $ & $82.1 \scriptsize{\raisebox{1pt}{$\pm 2.6$}} $ \\
           & LSTM & $94.2 \scriptsize{\raisebox{1pt}{$\pm 0.8$}}$ & $89.8 \scriptsize{\raisebox{1pt}{$\pm 1.0$}}$ & & $91.6 \scriptsize{\raisebox{1pt}{$\pm 2.2$}} $ & $89.5 \scriptsize{\raisebox{1pt}{$\pm 2.7$}} $ \\
           \midrule
          \multirow{3}{*}{Visual}
           & 1D-CNN & $67.3 \scriptsize{\raisebox{1pt}{$\pm 3.4$}}$ & $64.9 \scriptsize{\raisebox{1pt}{$\pm 2.9$}}$ & & $66.6 \scriptsize{\raisebox{1pt}{$\pm 3.1$}}$ & $61.6 \scriptsize{\raisebox{1pt}{$\pm 2.6$}}$ \\
           & LSTM & $87.5 \scriptsize{\raisebox{1pt}{$\pm 1.0$}}$ & $84.8 \scriptsize{\raisebox{1pt}{$\pm 0.9$}}$ & & $83.1 \scriptsize{\raisebox{1pt}{$\pm 0.8$}}$ & $80.7 \scriptsize{\raisebox{1pt}{$\pm 1.1$}}$ \\
           & Transformer & $88.3 \scriptsize{\raisebox{1pt}{$\pm 1.2$}}$ & $86.0 \scriptsize{\raisebox{1pt}{$\pm 1.0$}}$ & & $85.5 \scriptsize{\raisebox{1pt}{$\pm 0.9$}}$ & $82.1 \scriptsize{\raisebox{1pt}{$\pm 1.2$}}$ \\
          \bottomrule
    \end{tabular}

    \end{minipage}%
    \hspace{0.022\linewidth}%
    \begin{minipage}{.37\linewidth}
    \footnotesize
    \caption{Real-robot results evaluated for the tabletop scenario with different numbers of train or test objects present ($n_{\text{obj}}$).}\centering
    \label{tab:real_robot_results}
    \begin{tabular}{lccccc}
        \toprule
        \multirow{2}{*}{Model} & \multicolumn{2}{c}{Train ($n_{\text{obj}}$)} & & \multicolumn{2}{c}{Test ($n_{\text{obj}}$)} \\
        \cmidrule{2-3} \cmidrule{5-6} & 3 & 5 & & 3 & 5 \\
        \midrule
        LSTM & $6 / 10$ & $6 / 10$ & &  $6 / 10$ & $5 / 10$ \\
        Transformer & $4 / 10$ & $6 / 10$ & & $6 / 10$ & $5 /10$ \\
        \bottomrule
    \end{tabular}
    \end{minipage}
    \end{table*}

  \textbf{1D-CNN:}
  Temporal convolution has been successfully applied in prior work~\cite{Kumar2021} as a lightweight approach to capturing short-range dependencies.
  Here, we apply a three-layer 1D-CNN that convolves the feature representations across the time dimension to extract relevant temporal correlations.

  \textbf{LSTM:}
  The LSTM variant mirrors the recurrent teacher policy, consisting of a single-layer LSTM of size 768.

  \textbf{Transformer:}
  Given their proven effectiveness in capturing long-range dependencies, we evaluate the use of a transformer-based architecture for the student policy.
  This transformer comprises 3 layers, each with 8 heads and a hidden dimension of 256.


  \section{Experimental Setup}
  \subsection{Environments}
  We evaluate our method in simulated multi-object manipulation tasks using Nvidia Isaac Lab~\cite{Mittal2023}.
  The simulation consists of multiple parallel instances of our robotic system interacting with randomly selected YCB objects~\cite{Calli2015}.
  In total, we use 60 YCB objects, of which 48 are included in the training set, while 12 are held out to assess generalization.

  \subsection{Evaluation and Metrics}
  Performance is measured based on two success criteria: lifting the target object from the tabletop or bin and moving it to a specified 3D goal position, which is considered successful if the object is within 5\,cm of the target location.

  \section{Results}


  \subsection{Grasping from Tabletop Scenes}
  In cluttered environments, a proficient control policy should reliably grasp the target and move it to a desired position in 3D space.
  The  success rates for both criteria are presented in Table~\ref{tab:simulation_results}, showing that the LSTM teacher outperforms the MLP variant, with the best LSTM policy achieving a $94.2\%$ success rate averaged over all 48 training objects.

  To identify failure modes, we manually reviewed rollouts and found that most failures involved objects that were difficult to grasp, such as thin, elongated items like knives, which require precise handling to avoid excessive contact forces.
  Failures also occurred when smaller objects were covered by larger ones.
  While we observed some meaningful pre-grasp behaviors, such as reorienting objects or pushing obstacles aside, long-horizon strategies -- like intentionally moving a non-target object before retrieving the target -- remain difficult to learn.
  We depict the lifting success rate over the course of training on the left of Figure~\ref{fig:results}.
  Overall, we were able to learn highly effective teacher policies, that successfully handle diverse tabletop configurations.
  Failures are mainly limited to particularly difficult scenarios or slight exceedances of contact thresholds for hard-to-grasp items.

  In Table~\ref{tab:simulation_results}, we can see that the visual student policies are able to recover much of the teacher's performance.
  This indicates, that the formulation via supervised learning from a sequence of imperfect observations is an effective approach to transfer the policy's abilities to a real-world deployable observation space.
  On real-robot deployment (see Table~\ref{tab:real_robot_results}), we observed that the policies exhibited similar success rates on seen and unseen objects, indicating, that the learned behaviors are not overfitting to the training subset.
  
  \subsection{Grasping from Cluttered Bins}
  The previous results demonstrate that, for tabletop scenes, RL policies can learn to grasp objects from cluttered environments with high success rates.
  We aim to extend this capability to the more challenging scenario of picking objects from cluttered containers.
  This setup presents additional challenges, such as tightly packed heaps causing unforeseen interactions and the need for precise maneuvering to avoid collisions with the container walls.
  The second column of Figure~\ref{fig:results} shows the training progress for this scenario.

  The teacher policies for the bin-picking setup can learn careful grasping of diverse items, and underperform the tabletop policies only by a small margin.
  The learned policies again exhibit desirable behaviors like regrasping and pre-grasp manipulation.
  The policies cause more terminations due to contacts early in training, but converge to proficient and safe manipulation behaviors as training progresses.
  The student policies for this scenario again track the performance of the teachers closely.
  
  \subsection{Implicit Inference through Time}
  We hypothesize that the student infers the underlying object state necessary for imitating the teacher by leveraging the history of observations.
  If true, increasing the context length should enhance the student's ability to imitate the teacher's actions, resulting in lower loss.
  To investigate this, we log the loss curves of a history-aware student policy imitating an MLP teacher over different lengths of visible context in the bottom right of Figure~\ref{fig:results}.
  A clear pattern of decreasing loss with increasing context length is visible, showing that the student is better able to imitate the teacher when more context becomes available, which indicates that the student is indeed learning to infer the underlying object state from the history of observations.

  \section{Conclusion}
  \label{sec:conclusion}
  \textbf{Summary:} We have illustrated a way to condition RL policies on the output of SAM\,2 to achieve closed-loop, prompt-guided grasping from clutter.
  Specifically, we have formulated the problem of learning from imperfect detections of a foundation model as a POMDP and we have shown that it can be solved efficiently through history-aware architectures in a student-teacher setting.
  
  \textbf{Limitations and Future Work:} While we were able to create dexterous manipulation behaviors for cluttered bin-picking, we have left their real world deployment for future work.
  Further, while we have tackled the problem of picking and repositioning unknown objects, it would be interesting to apply our methodology to additional manipulation tasks.

  \section*{Acknowledgment}
  \footnotesize{This work has been funded by the German Ministry of Education and Research (BMBF), grant no. 01IS21080, project “Learn2Grasp: Learning Human-like Interactive Grasping based on Visual and Haptic Feedback”.}
  
  \clearpage
\bibliographystyle{IEEEtran}
\bibliography{references}


\end{document}